\documentclass[lettersize,journal]{IEEEtran}
\usepackage{amsmath,amsfonts}
\usepackage{algorithmic}
\usepackage{algorithm}
\usepackage{array}
\usepackage[caption=false,font=normalsize,labelfont=sf,textfont=sf]{subfig}
\usepackage{textcomp}
\usepackage{stfloats}
\usepackage{url}
\usepackage{verbatim}
\usepackage{graphicx}
\usepackage{cite}
\usepackage{booktabs}

\begin{document}

\title{Physics-Driven Learning Framework for Tomographic Tactile Sensing }

\author{Xuanxuan Yang, Xiuyang Zhang, Haofeng Chen, Gang Ma, Xiaojie Wang
        % <-this % stops a space
}

\markboth{Journal of \LaTeX\ Class Files,~Vol.~14, No.~8, August~2021}%
{Shell \MakeLowercase{\textit{et al.}}: A Sample Article Using IEEEtran.cls for IEEE Journals}



\maketitle

\begin{abstract}
Electrical impedance tomography (EIT) provides an attractive solution for large-area tactile sensing due to its minimal wiring and shape flexibility, but its nonlinear inverse problem often leads to severe artifacts and inaccurate contact reconstruction. This work presents PhyDNN, a physics-driven deep reconstruction framework that embeds the EIT forward model directly into the learning objective. By jointly minimizing the discrepancy between predicted and ground-truth conductivity maps and enforcing consistency with the forward PDE, PhyDNN reduces the black-box nature of deep networks and improves both physical plausibility and generalization. To enable efficient backpropagation, we design a differentiable forward-operator network that accurately approximates the nonlinear EIT response, allowing fast physics-guided training. Extensive simulations and real tactile experiments on a 16-electrode soft sensor show that PhyDNN consistently outperforms NOSER, TV, and standard DNNs in reconstructing contact shape, location, and pressure distribution. PhyDNN yields fewer artifacts, sharper boundaries, and higher metric scores, demonstrating its effectiveness for high-quality tomographic tactile sensing.

\end{abstract}


\section{Introduction}
\IEEEPARstart{H}{uman} skin exhibits exceptional pressure sensitivity and a wide dynamic range, enabling rich tactile interactions. Tactile sensors that emulate these capabilities have therefore become powerful tools for embodied intelligence and human–robot interaction. However, most conventional flexible tactile sensors are based on dense electrode arrays. When the number of sensing nodes increases, challenges such as complex wiring, increased hardware cost, and limited scalability arise.

Electrical Impedance Tomography (EIT)\cite{cheney1999electrical}–based tactile sensors have attracted growing attention due to their capability for large-area sensing with substantially reduced wiring. When external pressure is applied to the sensing material, its conductivity changes. By injecting currents through selected electrode pairs and measuring voltages on the remaining electrodes, an algorithm can reconstruct the material’s conductivity distribution, enabling the estimation of contact force magnitude and contact geometry. Furthermore, once an appropriate forward model is established, EIT theoretically enables tactile sensors of arbitrary shapes.
Traditional EIT-based tactile sensors typically rely on linear or iterative reconstruction algorithms such as NOSER\cite{cheney1990noser} and total variation (TV) regularization\cite{gonzalez2017isotropic}. However, EIT is an ill-posed and nonlinear inverse problem. These classical methods often introduce substantial artifacts, which degrade the accuracy of contact localization and shape identification. With the rise of deep learning, many studies have adopted fully data-driven end-to-end mappings from voltage measurements to conductivity images\cite{RN714} for contact force and shape estimation. Despite their convenience, purely black-box approaches may yield unpredictable behavior outside the training distribution.

To mitigate the black-box nature of deep neural networks and improve generalization and accuracy, recent research has explored embedding physical knowledge into the learning process. One line of work applies physics-informed preprocessing: a coarse conductivity image is first reconstructed using classical qualitative or quantitative EIT methods, and then a DNN is trained to refine it using ground-truth images as supervision. For example, Chen et al. proposed a physics-guided preprocessing method to improve force estimation in EIT-based tactile sensors\cite{chen2024enhancing}. However, the intermediate reconstruction may significantly deviate from the true conductivity distribution, since the DNN itself is trained without explicit physical supervision. Another line of research incorporates physical constraints directly into the network loss function, such as PDE-based physics-informed neural networks (PINNs)\cite{RN509,RN490,RN750} or Deep Image Prior (DIP)–based measurement-constrained methods\cite{liu2023deepeit}. Yet these approaches are essentially unsupervised and suffer from high computational cost, making them unsuitable for real-time tactile applications.

% Recently, Jin et al. introduced a physics-driven deep learning framework for addressing nonlinear geosteering inversion problems\cite{jin2020physics}. Inspired by their work, we propose a forward-model-embedded supervised learning framework for EIT-based tactile sensing. Unlike traditional machine learning approaches, our method incorporates not only the discrepancy between predicted and ground-truth conductivity, but also enforces consistency with the EIT forward model within the loss function. This reduces the black-box behavior, enhances physical interpretability, and improves both contact force estimation and contact shape reconstruction accuracy.

Recently, a growing body of research has explored integrating physical laws into deep learning to improve the reliability of inverse problem solutions. Jin et al.\cite{jin2020physics} introduced a physics-driven deep-learning framework for geosteering inversion, where a differentiable forward model and a CNN jointly enforce consistency between predictions and electromagnetic measurements. Their results demonstrated that embedding the forward model into the loss function yields solutions that are more stable and physically meaningful than purely data-driven mappings. In structural engineering, Zhang et al.\cite{zhang2020physics} proposed a physics-guided CNN for seismic response prediction, showing that incorporating dynamics-based constraints alleviates overfitting, reduces the need for large datasets, and significantly improves robustness. In photonic device design, Jiang et al.\cite{jiang2019global} further demonstrated the power of physics-informed learning by embedding adjoint electromagnetic simulations directly into a generative neural optimizer, enabling efficient surrogate modeling without requiring large precomputed datasets. Inspired by these advances, we introduce a physics-driven, forward-model-embedded supervised learning framework for EIT-based tactile sensing. Unlike conventional deep learning approaches that rely solely on measurement-to-image mappings, our method jointly minimizes the discrepancy between predicted and ground-truth conductivity while enforcing consistency with the EIT forward model. By integrating differentiable physics into the learning process, the proposed framework reduces black-box behavior, enhances physical interpretability, and significantly improves both shape reconstruction and force-distribution estimation in tactile sensing.

Our contributions are summarized as follows:
\begin{enumerate}
\item We propose a physics-driven supervised learning framework for EIT-based tactile sensing, which integrates the forward model into the loss function to improve physical consistency and reduce the black-box nature of deep networks.

\item We demonstrate that incorporating forward-model constraints significantly enhances contact-force estimation and contact-shape reconstruction, outperforming purely data-driven approaches in accuracy and generalization.

\item We design an efficient differentiable forward-operator strategy, enabling fast training and real-time inference while maintaining strong physical interpretability.
\end{enumerate}

\section{Methods and System Design}
\subsection{Tactile sensor fabrication and sensing system }
\begin{figure*}[!t]
\centering
\includegraphics[]{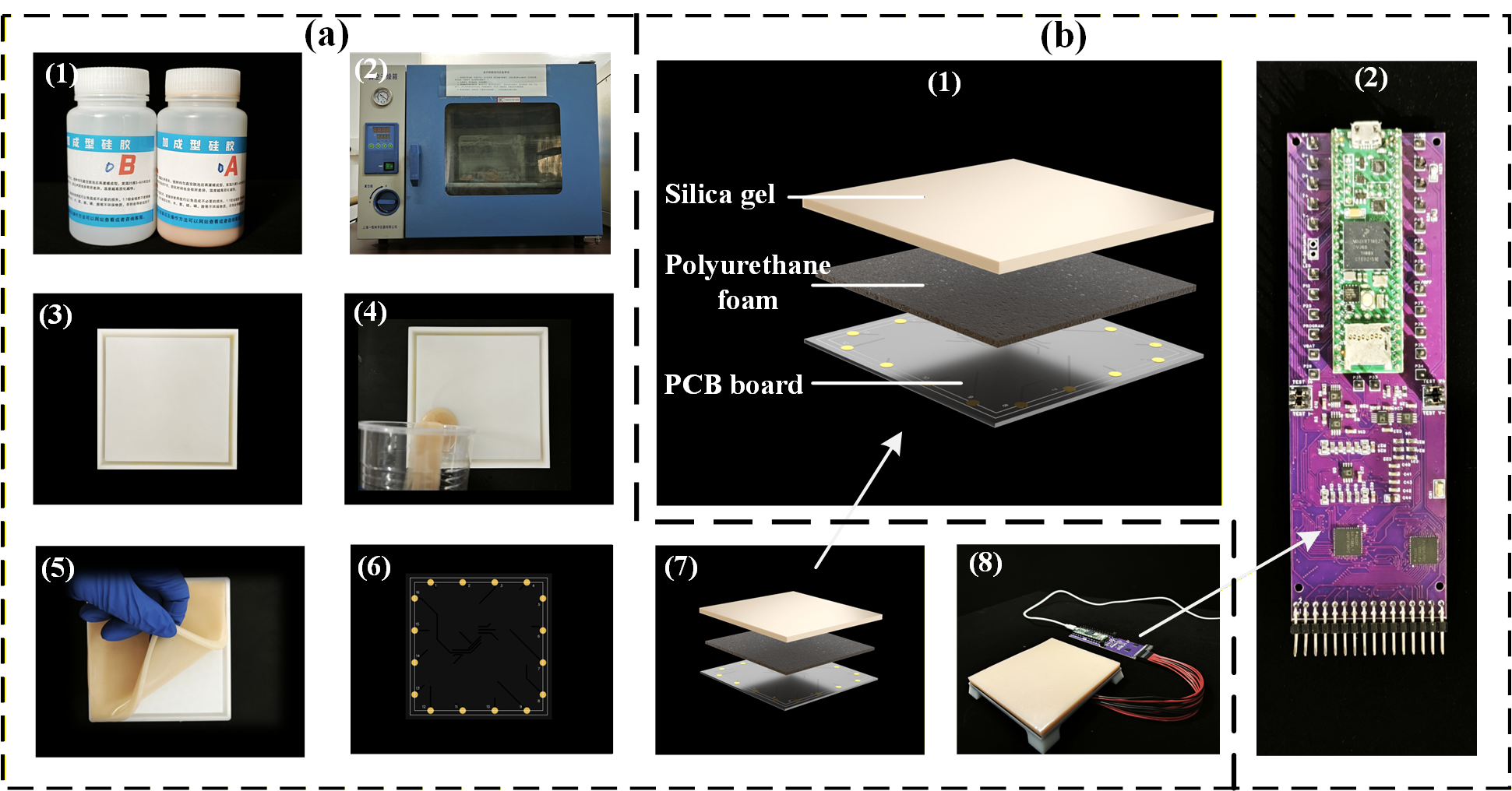}
\caption{Fabrication process and sensing system of the proposed EIT-based tactile sensor. 
(a) Step-by-step fabrication of the biomimetic multilayer tactile structure, including silicone insulation preparation, mold casting, bubble removal, curing, and integration with the water-infused polyurethane foam layer. 
(b) Final assembled tactile sensor and the customized signal acquisition circuit based on a Teensy~4.1 controller with optimized PCB routing and impedance-matching design.}
\label{flowchart}
\end{figure*}

Fig. \ref{flowchart}(a)(1)–(8) illustrates the fabrication process of our tactile sensor. To prepare the biomimetic silicone insulating layer, we used skin-tone silicone (hardness 0, DePing Co., China). Components A and B were mixed at a 1:1 ratio and thoroughly stirred. The mixture was then placed in a DZF-6050 vacuum drying oven (Shanghai Yiheng Scientific Instruments Co., Ltd.) for 2 minutes to remove air bubbles. After degassing, the silicone was cast into a 3D-printed mold, cured for 24 hours at room temperature, and subsequently demolded. Next, a soft polyurethane foam infused with tap water was positioned on the custom-designed PCB, which contains 16 uniformly distributed electrodes. The silicone insulating layer was then placed on top and sealed to form the final multilayer structure, as shown in Fig. \ref{flowchart}(b)(1). Fig. \ref{flowchart}(b)(2) shows the signal acquisition circuit, which is an improved version based on the design of Zhu et al.\cite{zhu2021eit}. We adopted a Teensy 4.1 microcontroller as the main control unit and redesigned the PCB layout to reduce wiring complexity and enhance noise immunity. To match the impedance characteristics of our tactile sensor, the resistance values of R7, R8, R21, and R22 were adjusted to 1 $\mathrm{k}\Omega $. The excitation frequency was set to 50 kHz, and the system achieves a real-time imaging rate of 10 frames per second.

\subsection{Proposed Framework: PhyDNN}
\begin{figure}[htbp]
\centering
\includegraphics[scale=1]{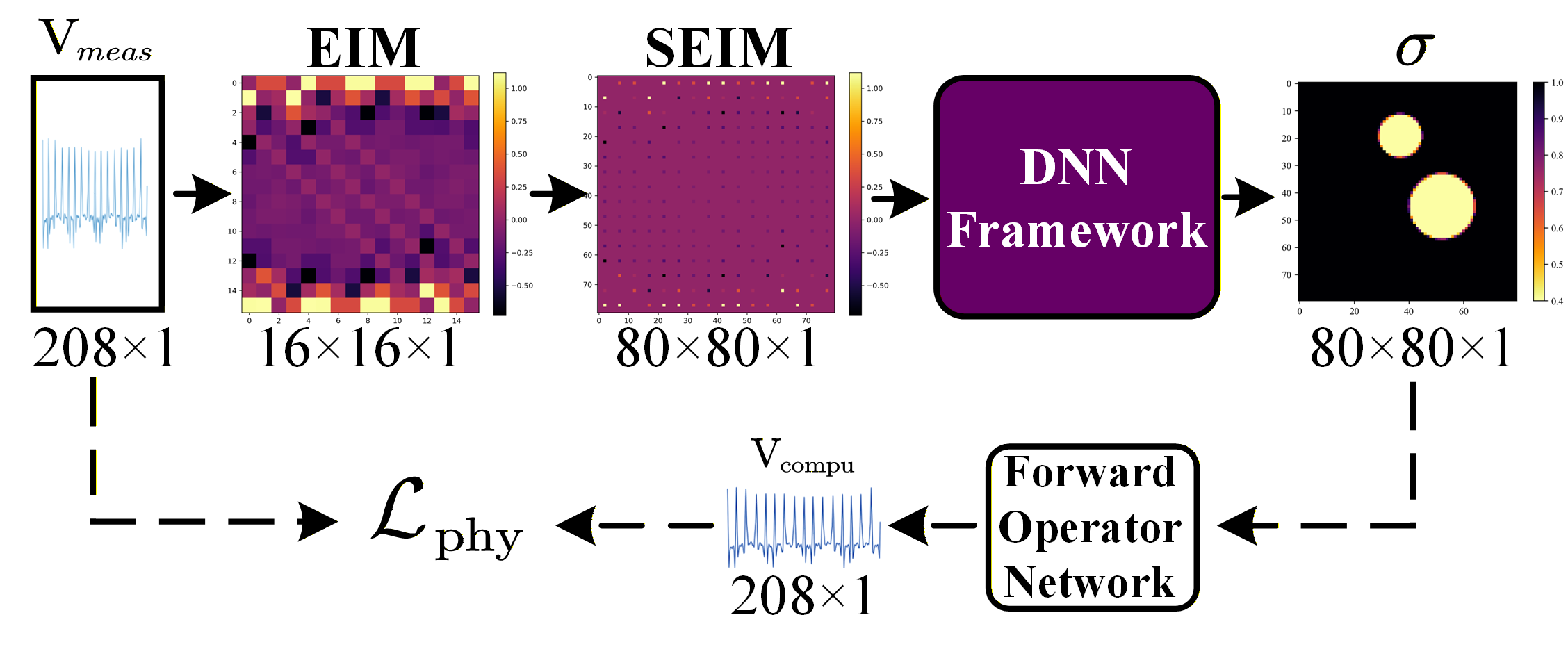}
\caption{Overall architecture of the proposed physics-driven neural network (PhyDNN) .}
\label{phydnn}
\end{figure}

In this work, we refer to our physics-driven reconstruction framework as PhyDNN.  
The overall architecture is shown in Fig. \ref{phydnn}. The model takes the measured voltage vector 
$\mathbf{V}_{\mathrm{meas}}$ as input and outputs the conductivity distribution 
$\sigma \in \mathbb{R}^{80\times 80}$.  
The framework is flexible and allows different backbone networks; in this paper,
we adopt the classical U-Net\cite{ronneberger2015u} due to its strong capability in dense prediction tasks.

The training objective of PhyDNN is defined as:
\begin{equation}
\mathcal{L}
= \alpha \mathcal{L}_{\mathrm{data}}
+ \beta \mathcal{L}_{\mathrm{phy}}
= \alpha \lVert \sigma - \sigma_p \rVert_2^2
+ \beta \lVert F(\sigma_p) - \mathbf{V}_{\mathrm{meas}} \rVert_2^2 ,
\end{equation}
where $\alpha$ and $\beta$ are weighting coefficients, 
$\sigma_p$ is the predicted conductivity, and
$F(\cdot)$ denotes the EIT forward operator.
When $\alpha>0$, the framework performs \emph{supervised learning} using
ground-truth conductivity maps; when $\alpha=0$, the loss reduces to $\mathcal{L} = \lVert F(\sigma_p) - \mathbf{V}_{\mathrm{meas}} \rVert_2^2 $, which corresponds to the unsupervised setting adopted by DeepEIT~\cite{liu2023deepeit}.
Since unsupervised schemes lack real-time capability, we focus on the 
supervised regime ($\alpha > 0$) in this work.

\subsection{Forward-Operator Network}

The EIT forward problem is governed by the elliptic PDE
\begin{equation}
\nabla \!\cdot\! \left( \sigma \nabla u \right) = 0 , \quad \text{in } \Omega ,
\label{eq:pde}
\end{equation}
where $u$ denotes the electrical potential.  
Incorporating the physical loss $\mathcal{L}_{\mathrm{phy}} = 
\lVert F(\sigma_p) - \mathbf{V}_{\mathrm{meas}} \rVert_2^2$ requires evaluating
the forward operator $F(\cdot)$ and its gradient with respect to $\sigma_p$.
However, classical numerical solvers such as \texttt{EIDORS}~\cite{RN83} and 
\texttt{pyEIT}~\cite{liu2018pyeit} are not differentiable and computing their Fréchet
derivative $\partial F / \partial \sigma$ is computationally prohibitive for
end-to-end deep learning.

\begin{figure}[htbp]
\centering
\includegraphics[scale=1]{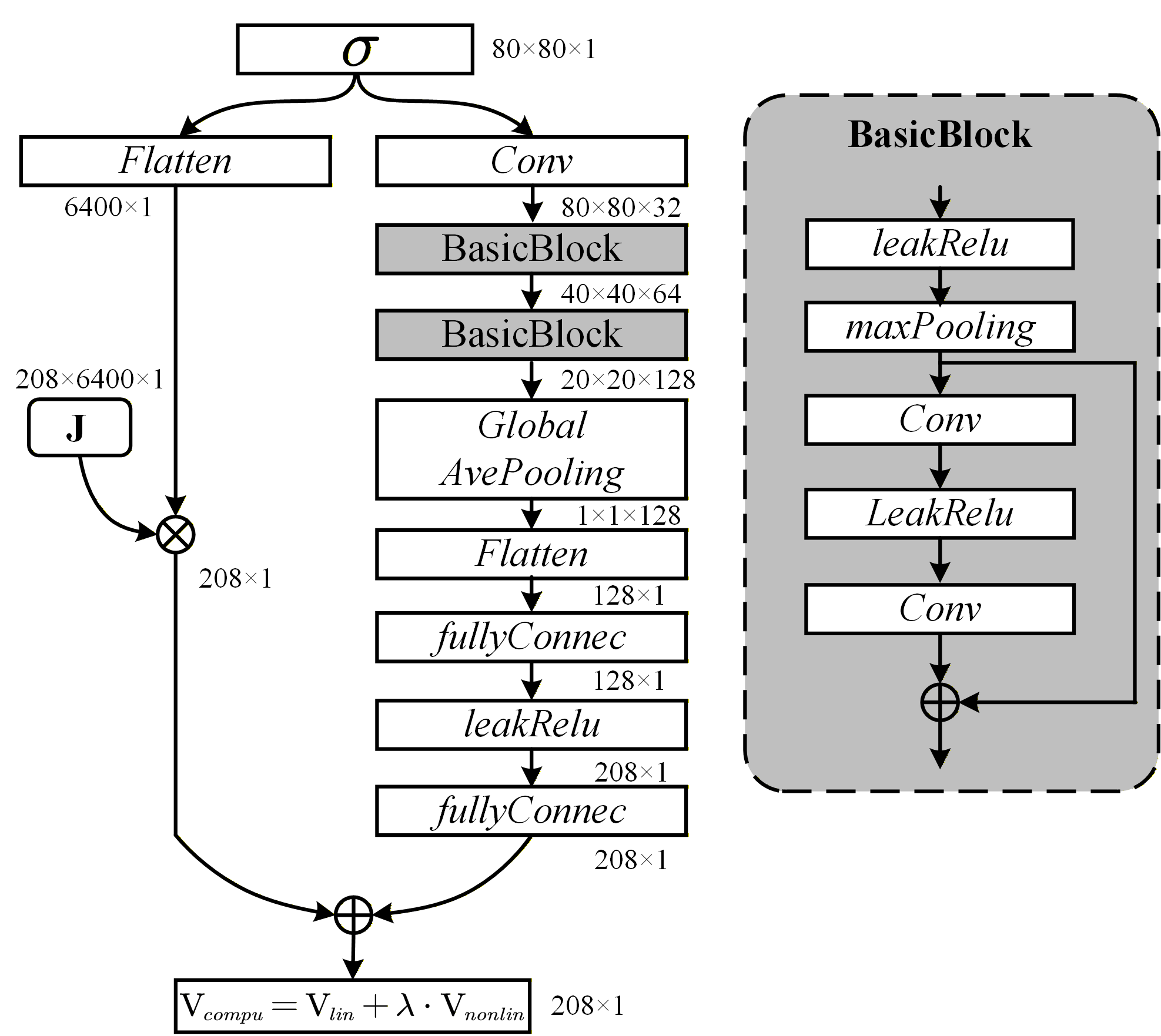}
\caption{Structure of the CNN-based forward-operator network. }
\label{fornet}
\end{figure}

To enable efficient training, we introduce a differentiable CNN-based 
forward-operator network (Fig.~\ref{fornet}) that approximates the mapping 
\begin{equation}
F:\sigma \in \mathbb{R} ^{80\times 80}\,\,\longmapsto \,\,\mathrm{V}\in \mathbb{R} ^{208}
\end{equation}

Once learned, this surrogate forward model is fully compatible with automatic
differentiation frameworks (e.g., PyTorch), allowing the physical loss to be
optimized jointly with the reconstruction network.

The proposed operator network consists of three components:
(1) a linear physics branch implementing the sensitivity matrix $J$;
(2) a nonlinear CNN correction branch that compensates for linear-model errors; and
(3) a learnable fusion coefficient that balances the two outputs.
Together, these modules provide an accurate and computationally efficient surrogate
of the EIT forward operator, enabling fast, physics-aware training of PhyDNN.

\section{Simulation Studies}

\subsection{Dataset}

We constructed a finite element model (FEM) of the EIT-based tactile sensor using the EIDORS toolbox. The model employs 16 electrodes and adopts the 3--3 excitation and measurement pattern \cite{smela2023eit}. The FEM consists of 6400 elements (80$\times$80). A total of 24,000 training samples were generated, containing several representative shapes, including single circles, double circles, L-shaped objects, isosceles triangles, rectangles, and concentric rings. The positions and sizes of the inclusions were randomized, and their conductivity values were uniformly sampled between 0.1 and 0.9~S/m. In addition, 6,000 testing samples were generated following the same procedure. To better simulate the noise encountered in practical measurements, data augmentation was applied: Gaussian white noise was added to half of the training samples, resulting in a random signal-to-noise ratio (SNR) between 30~dB and 60~dB. This strategy improves the robustness and generalization ability of the model.

% \subsection{Training and Evaluation of the Forward Operator Network}
% Before training PhyDNN, we first train the proposed CNN-based forward-operator network. The network is trained for 100 epochs on an NVIDIA GTX3090 using the mean squared error between predicted and simulated voltages as the objective. To evaluate its accuracy, we compare the network output against voltages generated by the classical linear Jacobian operator~$J$, a standard approximation used in traditional EIT solvers. A randomly selected example is shown in Fig.~\ref{voltage}, where the proposed operator provides a visibly closer match to the true voltages than $J$. We further assess performance over all 6000 test samples by computing the average prediction error and correlation coefficient. The results (Table~\ref{tab:forward_operator}) show that the learned operator substantially improves both metrics compared with the Jacobian-based model. This enhanced fidelity is critical for PhyDNN, as it strengthens the physical consistency enforced by the physics-driven loss during end-to-end training.

\subsection{Training and Evaluation of the Forward Operator Network}

\begin{figure}[htbp]
\centering
\includegraphics[]{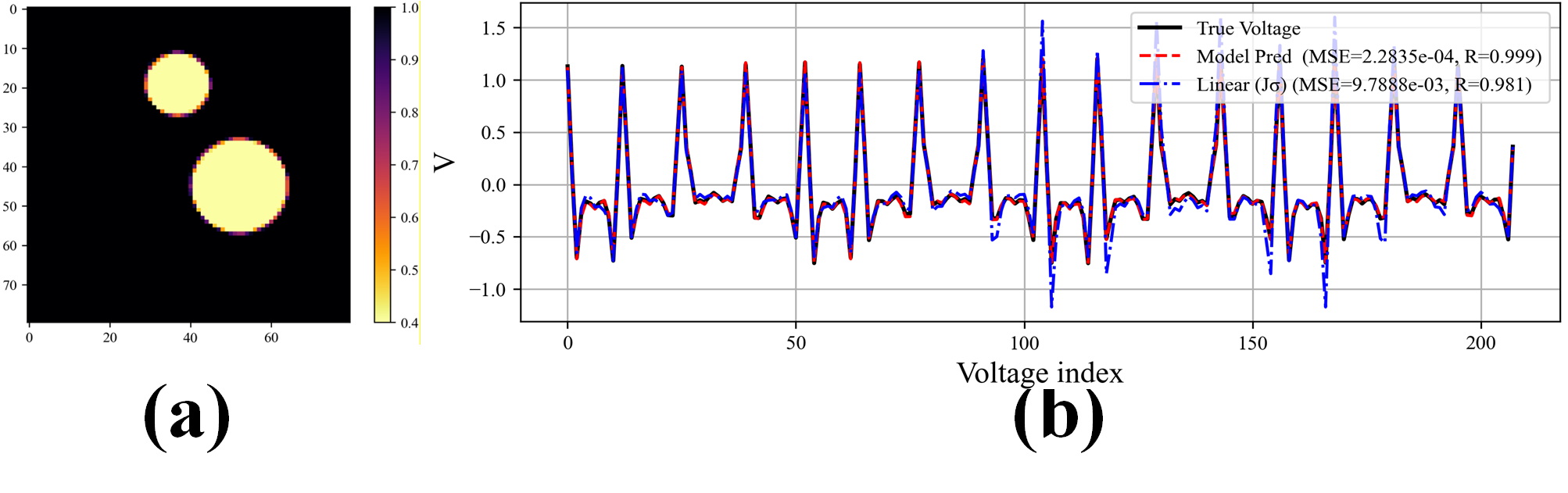}
\caption{Evaluation of the learned forward operator. 
(a) Randomly selected conductivity distribution from the test set. 
(b) Comparison of boundary voltages predicted by the learned CNN-based operator and by the linear Jacobian operator~$J$. 
The learned operator exhibits higher fidelity to the true simulated voltages, enabling more accurate physics-based supervision in PhyDNN.}
\label{voltage}
\end{figure}

Before training PhyDNN, we first train the proposed CNN-based forward-operator network, which serves as a differentiable surrogate for the nonlinear EIT forward mapping. The network is trained for 100 epochs on an NVIDIA GTX\,3090 using the mean squared error (MSE) between predicted and reference voltages as the loss function. Here, MSE measures the average squared difference between two voltage vectors and quantifies the accuracy of the forward prediction. To assess the fidelity of the learned operator, we compare its voltage predictions with those generated by the classical linear Jacobian operator~$J$, which is widely used as a first-order approximation in traditional EIT solvers. A representative test example is shown in Fig.~\ref{voltage}. Fig.~\ref{voltage}(a) displays a randomly selected conductivity sample from the test set, while Fig.~\ref{voltage}(b) visualizes the corresponding boundary voltages predicted by the learned forward operator and by $J$. The CNN-based operator clearly produces a closer match to the true voltages, exhibiting reduced error and improved structural similarity. To evaluate performance systematically, we compute the average MSE and correlation coefficient across all 6000 test samples. As summarized in Table~\ref{tab:forward_operator}, the learned operator achieves a significantly lower MSE and a higher correlation value than the Jacobian-based model. These results demonstrate that the differentiable operator provides a more accurate approximation of the underlying EIT physics, thereby strengthening the reliability of the physics-driven loss when integrated into PhyDNN training.

\begin{table}[htbp]
\centering
\caption{Performance comparison between the learned forward operator and the linear Jacobian model over 6000 test samples}
\begin{tabular}{c c c}
\toprule
Method & MSE $\downarrow$ & Correlation $R$ $\uparrow$ \\
\midrule
Learned forward operator & $2.57\times10^{-3}$ & $0.99498$ \\
Linear Jacobian ($J$)   & $1.62\times10^{-2}$ & $0.97123$ \\
\bottomrule
\end{tabular}
\label{tab:forward_operator}
\end{table}

\begin{figure*}[htbp]
\centering
\includegraphics[scale=1]{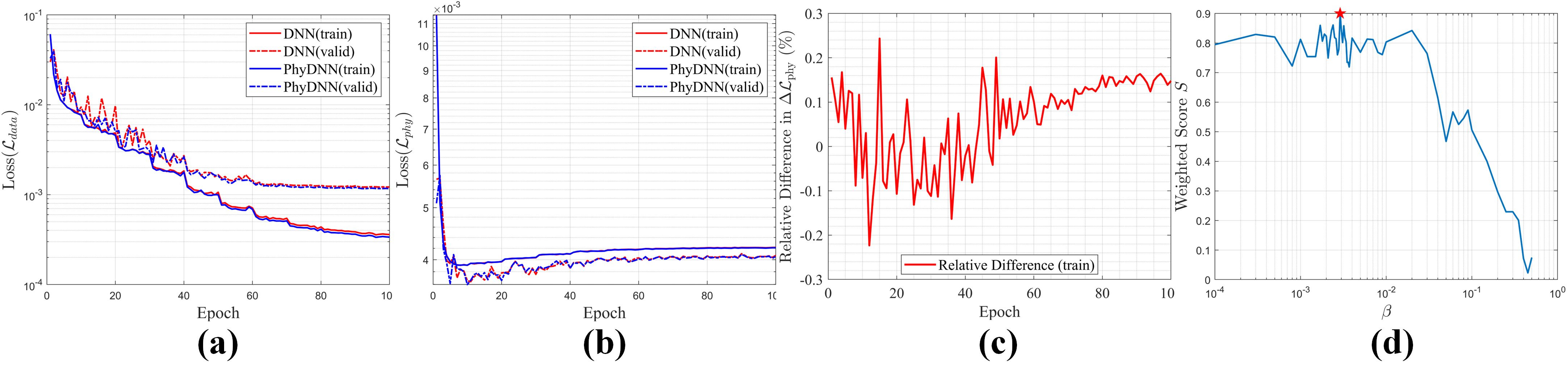}
\caption{Training and evaluation of the PhyDNN framework. 
(a) Training and validation data loss $\mathcal{L}_{\mathrm{data}}$ for DNN and PhyDNN. 
(b) Training and validation physical loss $\mathcal{L}_{\mathrm{phy}}$. 
(c) Relative difference between the physical losses of DNN and PhyDNN, highlighting the effect of the physics constraint after the warm-up stage. 
(d) Grid-search results of the weighted reconstruction score for 52 values of $\beta$, showing that $\beta = 0.0029$ yields the best overall performance.}

\label{loss}
\end{figure*}

\subsection{Training and Evaluation of the PhyDNN}

To assess the contribution of the physics-based loss, we conduct an ablation study with $\alpha=1$ and two settings: a purely data-driven DNN ($\beta=0$) and the proposed PhyDNN ($\beta=0.0029$). Both models are trained for 100 epochs under identical conditions.
A 30-epoch warm-up strategy is applied, during which the physical loss is computed but excluded from backpropagation. This prevents unstable early predictions from adversely influencing the physics constraint. After warm-up, the physical term is activated only for PhyDNN, while for the baseline DNN it is monitored but never optimized.
Fig.~\ref{loss}(a) presents the data loss curves. PhyDNN consistently achieves lower training and validation errors, indicating that enforcing forward-model consistency enhances the quality of supervised learning. Fig.~\ref{loss}(b) shows the physical loss. Although the absolute gap between the two models appears small—both converge to similar magnitudes due to the strong structure imposed by the measurement pattern—PhyDNN maintains a consistently lower value once the physical constraint becomes active.
To highlight this improvement, Fig.~\ref{loss}(c) plots the relative difference
$\Delta\mathcal{L}_{\mathrm{phy}}
=
\left(
\mathcal{L}^{\mathrm{DNN}}_{\mathrm{phy}}
-
\mathcal{L}^{\mathrm{PhyDNN}}_{\mathrm{phy}}
\right)
/\mathcal{L}^{\mathrm{DNN}}_{\mathrm{phy}}
\times 100\%.$
During warm-up, the two curves occasionally cross because neither model is constrained by the PDE. After activation, PhyDNN quickly stabilizes and consistently outperforms the baseline. This confirms that integrating the forward operator effectively steers the network toward physically valid conductivity distributions.

Having verified the benefit of the physical term, we next perform a grid search to determine the optimal weighting coefficient $\beta$.

\begin{figure*}[htbp]
\centering
\includegraphics[]{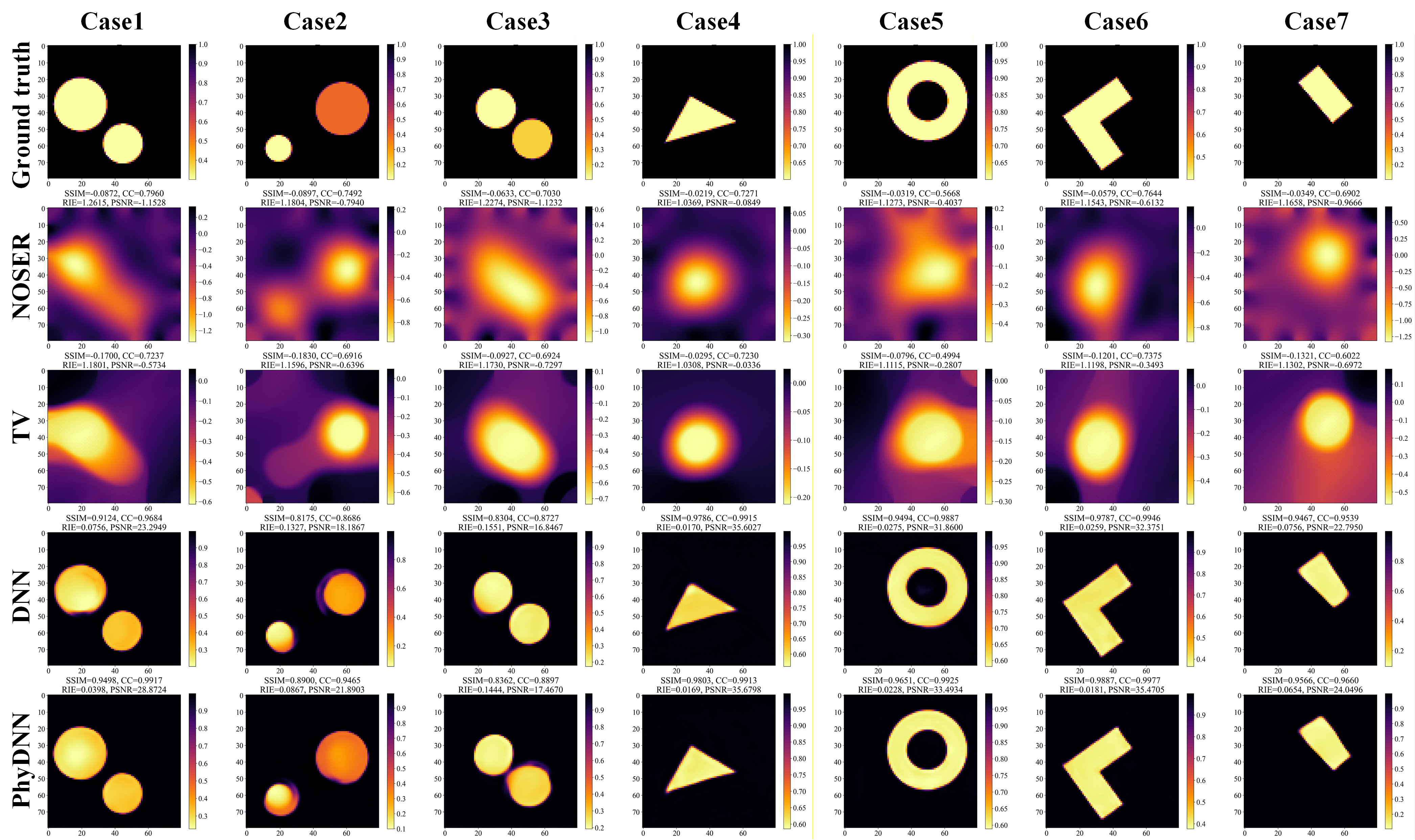}
\caption{Simulation reconstruction results using NOSER, TV, DNN, and the proposed PhyDNN. }

\label{simu}
\end{figure*}

\begin{table}[htbp]
\centering
\caption{Average quantitative metrics over 6000 simulated test samples}
\begin{tabular}{c c c c c}
\toprule
Metric & NOSER & TV & DNN & PhyDNN \\
\midrule
SSIM $\uparrow$ & -0.0127 & -0.3570 & 0.9754 & 0.9765 \\
CC $\uparrow$  & 0.1207  & 0.0696  & 0.9875 & 0.9880 \\
RIE $\downarrow$  & 1.2130  & 1.5661  & 0.0295 & 0.0290 \\
PSNR (dB) $\uparrow$ & -1.0795 & -3.3490 & 33.1230 & 33.2950 \\
\bottomrule
\end{tabular}
\label{tab:sim_metrics}
\end{table}

\subsection{Grid Search for the Physical Weight}
The choice of the physical weight $\beta$ plays a critical role in PhyDNN. 
If $\beta$ is too large, the training process overemphasizes the forward PDE constraint, 
which may suppress the recovery of high-resolution image details. 
Conversely, if $\beta$ is too small, the physical constraint becomes insufficient, 
and the model behaves similarly to a purely data-driven DNN.

To determine a suitable value, we perform a grid search over 52 values of $\beta$ 
ranging from $10^{-4}$ to $0.5$.  
For each value, we train the model using identical settings and evaluate its reconstruction 
quality on all 6000 testing samples.  
Four widely used EIT metrics are employed: structural similarity(SSIM), correlation coefficient (CC), relative image error (RIE), 
and peak signal to noise ratio (PSNR).  
To provide an overall performance indicator, we compute a weighted score defined as $S = 0.25\,(\mathrm{SSIM} + \mathrm{CC} + \mathrm{PSNR}) 
      - 0.25\,\mathrm{RIE}$, where higher $S$ indicates better reconstruction performance.

The averaged weighted score for all 52 values is shown in Fig.~\ref{loss}(d).  
The peak occurs at $\beta = 0.0029$, indicating that this value provides the best balance between 
physical regularization and data fidelity, and therefore yields the optimal enhancement from the 
physics-based loss.

\subsection{Simulation result}

To evaluate the reconstruction performance of the proposed PhyDNN framework, we conducted a series of simulation experiments using unseen test samples. Seven representative cases were randomly selected from different geometric categories in the test set, including single and double circular inclusions, triangular shapes, concentric rings, rectangular objects, and L-shaped targets. Four reconstruction methods were compared: NOSER , TV , the baseline DNN, and the proposed PhyDNN. For each reconstruction, we computed four quantitative metrics—SSIM, CC, RIE, and PSNR—and displayed them beneath the corresponding images, as shown in Fig.~\ref{simu}. For the double-inclusion cases (Cases 1--3), NOSER and TV exhibit strong artifacts and insufficient resolution, making it difficult to distinguish the two circular objects. In contrast, the learning-based DNN and PhyDNN produce significantly cleaner reconstructions with far fewer artifacts. Moreover, PhyDNN provides more accurate shape recovery than the baseline DNN, highlighting the benefit of the physics-guided loss, which constrains the solution toward physically valid conductivity distributions. For the more complex shapes (Cases 4--7), including triangles, concentric rings, rectangles, and L-shaped inclusions, NOSER and TV struggle to produce reliable images and suffer from severe distortions. The DNN and PhyDNN, however, achieve markedly improved reconstruction quality. In particular, PhyDNN demonstrates superior shape fidelity in Cases 4, 5, and 7, producing results that closely match the ground truth. Although the difference is visually subtle for the L-shaped object (Case~6), PhyDNN still yields a more accurate conductivity estimation compared to the baseline. For all cases, PhyDNN consistently achieves the highest scores across all four quantitative metrics.

In addition to the qualitative visual comparison, we further computed the average SSIM, CC, RIE, and PSNR scores across all 6000 test samples for the four methods. The results, summarized in Table~\ref{tab:sim_metrics}, confirm that PhyDNN achieves the best overall performance, demonstrating its strong robustness and generalization capability.

\section{Experimental Studies}

\begin{figure*}[htbp]
\centering
\includegraphics[]{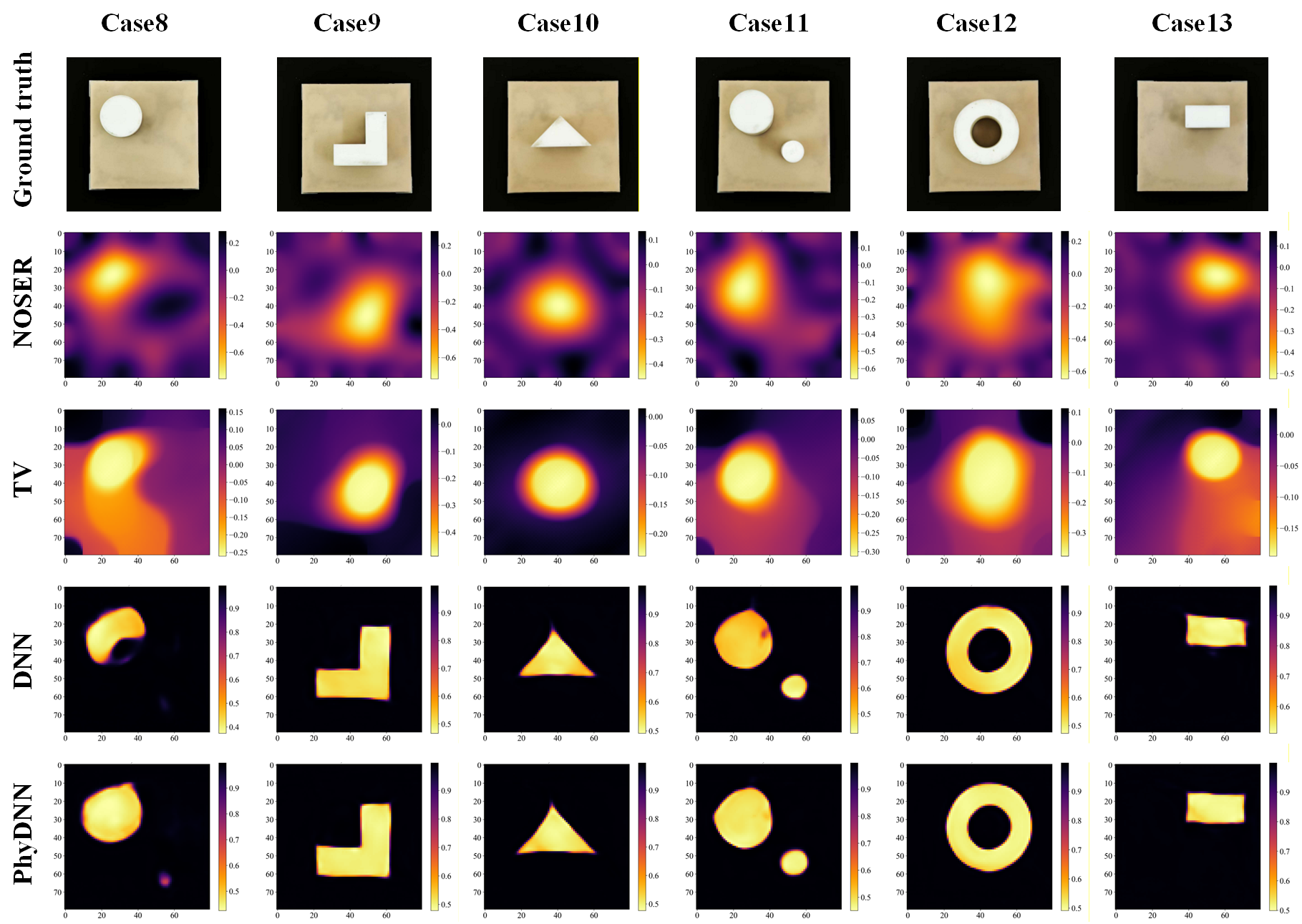}
\caption{Real-world tactile reconstructions obtained from the fabricated 10cm $\times $ 10cm EIT-based sensor. 
}

\label{experi}
\end{figure*}

To further assess the advantages of the proposed PhyDNN framework, we conducted real-world experiments using the $10\,\mathrm{cm}$ tactile sensor fabricated as described in this paper. Six representative objects were used to perform uniform pressing tests, including circles with diameters of $4\,\mathrm{cm}$ and $2\,\mathrm{cm}$, a right isosceles triangle with leg length $4\,\mathrm{cm}$, a concentric ring (outer diameter $6\,\mathrm{cm}$, inner diameter $3\,\mathrm{cm}$), an L-shaped object ($5\,\mathrm{cm}$ long, $2\,\mathrm{cm}$ wide), and a rectangle. The reconstructed images obtained using NOSER, TV, DNN, and PhyDNN are shown in Fig.~\ref{experi}. Overall, NOSER and TV produce reconstructions with significant artifacts and blurred boundaries, making it difficult to correctly identify object shapes or accurately locate the pressed region. In contrast, the data-driven DNN demonstrates better shape and position reconstruction, producing clearer outlines and improved localization. PhyDNN further enhances reconstruction quality across all test cases. For case 8, PhyDNN produces a noticeably more circular and sharper reconstruction than the baseline DNN. For cases 9 and 10, the conductivity distributions reconstructed by PhyDNN are more uniform and better aligned with the true geometry. For the more complex case~11 (L-shape), PhyDNN yields fewer artifacts and more accurately recovers the concave structure. In case~12 (concentric ring), the ring boundaries are more complete and circular under PhyDNN. In case~13 (rectangle), PhyDNN reconstructs straighter edges and a more faithful overall geometry.

These results demonstrate that incorporating the physical loss significantly improves the network’s robustness and generalization in real measurement conditions. By constraining the reconstructed conductivity to be consistent with the governing forward model, PhyDNN mitigates artifacts, enhances shape fidelity, and provides more reliable spatial localization compared with both classical methods and purely data-driven neural networks.

\section{Conclusion}
We presented PhyDNN, a physics-driven reconstruction framework for EIT-based tactile sensing. By embedding a differentiable forward model into the loss function, PhyDNN enforces physical consistency during learning, reducing artifacts and improving the fidelity of contact-shape and contact-location estimation. The proposed forward-operator network enables efficient end-to-end training with automatic differentiation, making the method suitable for real-time tactile applications. Both simulation studies and real-world experiments demonstrate that PhyDNN substantially outperforms classical EIT solvers and purely data-driven DNNs. The method yields more accurate conductivity distributions, sharper geometric features, and improved robustness to noise. These results indicate that physics-driven learning is a promising direction for next-generation tomographic tactile sensors, enabling reliable, high-resolution haptic perception with minimal wiring and scalable sensor designs.

\bibliographystyle{IEEEtran}
\bibliography{ref.bib}

@article{cheney1999electrical,
  title={Electrical impedance tomography},
  author={Cheney, Margaret and Isaacson, David and Newell, Jonathan C},
  journal={SIAM review},
  volume={41},
  number={1},
  pages={85--101},
  year={1999},
  publisher={SIAM}
}

@article{cheney1990noser,
  title={NOSER: An algorithm for solving the inverse conductivity problem},
  author={Cheney, Margaret and Isaacson, David and Newell, Jonathan C and Simske, S and Goble, J},
  journal={International Journal of Imaging systems and technology},
  volume={2},
  number={2},
  pages={66--75},
  year={1990},
  publisher={Wiley Online Library}
}

@article{gonzalez2017isotropic,
  title={Isotropic and anisotropic total variation regularization in electrical impedance tomography},
  author={Gonz{\'a}lez, Gerardo and Kolehmainen, Ville and Sepp{\"a}nen, Aku},
  journal={Computers \& Mathematics with Applications},
  volume={74},
  number={3},
  pages={564--576},
  year={2017},
  publisher={Elsevier}
}

@article{liu2023deepeit,
  title={DeepEIT: Deep image prior enabled electrical impedance tomography},
  author={Liu, Dong and Wang, Junwu and Shan, Qianxue and Smyl, Danny and Deng, Jiansong and Du, Jiangfeng},
  journal={IEEE Transactions on Pattern Analysis and Machine Intelligence},
  volume={45},
  number={8},
  pages={9627--9638},
  year={2023},
  publisher={IEEE}
}

@article{RN509,
   author = {Bar, Leah and Sochen, Nir},
   title = {Strong Solutions for PDE-Based Tomography by Unsupervised Learning},
   journal = {SIAM Journal on Imaging Sciences},
   volume = {14},
   number = {1},
   pages = {128-155},
   DOI = {10.1137/20m1332827},
   url = {https://epubs.siam.org/doi/abs/10.1137/20M1332827},
   year = {2021},
   type = {Journal Article}
}

@inproceedings{RN490,
   author = {Pokkunuru, Akarsh and Rooshenas, Pedram and Strauss, Thilo and Abhishek, Anuj and Khan, Taufiquar},
   title = {Improved Training of Physics-Informed Neural Networks Using Energy-Based Priors: a Study on Electrical Impedance Tomography},
   booktitle = {The Eleventh International Conference on Learning Representations},
   type = {Conference Proceedings}
}

@article{RN750,
   author = {Yang, X. and Zhang, Y. and Chen, H. and Ma, G. and Wang, X.},
   title = {A Two-Stage Imaging Framework Combining CNN and Physics-Informed Neural Networks for Full-Inverse Tomography: A Case Study in Electrical Impedance Tomography (EIT)},
   journal = {IEEE Signal Processing Letters},
   pages = {1-5},
   ISSN = {1558-2361},
   DOI = {10.1109/LSP.2025.3545306},
   year = {2025},
   type = {Journal Article}
}

@inproceedings{chen2024enhancing,
  title={Enhancing Tactile Sensing in Robotics: Dual-Modal Force and Shape Perception with EIT-based Sensors and MM-CNN},
  author={Chen, Haofeng and Yang, Xuanxuan and Ma, Gang and Wang, Yucheng and Wang, Xiaojie},
  booktitle={2024 IEEE International Conference on Robotics and Automation (ICRA)},
  pages={3311--3317},
  year={2024},
  organization={IEEE}
}

@article{jin2020physics,
  title={A physics-driven deep-learning network for solving nonlinear inverse problems},
  author={Jin, Yuchen and Shen, Qiuyang and Wu, Xuqing and Chen, Jiefu and Huang, Yueqin},
  journal={Petrophysics},
  volume={61},
  number={01},
  pages={86--98},
  year={2020},
  publisher={SPWLA}
}

@inproceedings{zhu2021eit,
  title={EIT-kit: An electrical impedance tomography toolkit for health and motion sensing},
  author={Zhu, Junyi and Snowden, Jackson C and Verdejo, Joshua and Chen, Emily and Zhang, Paul and Ghaednia, Hamid and Schwab, Joseph H and Mueller, Stefanie},
  booktitle={The 34th Annual ACM Symposium on User Interface Software and Technology},
  pages={400--413},
  year={2021}
}

@inproceedings{ronneberger2015u,
  title={U-net: Convolutional networks for biomedical image segmentation},
  author={Ronneberger, Olaf and Fischer, Philipp and Brox, Thomas},
  booktitle={International Conference on Medical image computing and computer-assisted intervention},
  pages={234--241},
  year={2015},
  organization={Springer}
}

@article{smela2023eit,
  title={EIT for tactile sensing: considerations regarding the injection-measurement pattern},
  author={Smela, Elisabeth},
  journal={Engineering Research Express},
  volume={4},
  number={4},
  pages={045041},
  year={2023},
  publisher={IOP Publishing}
}

@inproceedings{RN714,
   author = {Li, Xiuyan and Lu, Yang and Wang, Jianming and Dang, Xin and Wang, Qi and Duan, Xiaojie and Sun, Yukuan},
   title = {An image reconstruction framework based on deep neural network for electrical impedance tomography},
   booktitle = {2017 IEEE International Conference on Image Processing (ICIP)},
   publisher = {IEEE},
   pages = {3585-3589},
   ISBN = {1509021752},
   type = {Conference Proceedings}
}

@article{liu2018pyeit,
  title={pyEIT: A python based framework for Electrical Impedance Tomography},
  author={Liu, Benyuan and Yang, Bin and Xu, Canhua and Xia, Junying and Dai, Meng and Ji, Zhenyu and You, Fusheng and Dong, Xiuzhen and Shi, Xuetao and Fu, Feng},
  journal={SoftwareX},
  volume={7},
  pages={304--308},
  year={2018},
  publisher={Elsevier}
}

@article{RN83,
   author = {Adler, Andy and Lionheart, William},
   title = {EIDORS: Towards a community-based extensible software base for EIT},
   year = {2005},
   type = {Journal Article}
}

@article{jiang2019global,
  title={Global optimization of dielectric metasurfaces using a physics-driven neural network},
  author={Jiang, Jiaqi and Fan, Jonathan A},
  journal={Nano letters},
  volume={19},
  number={8},
  pages={5366--5372},
  year={2019},
  publisher={ACS Publications}
}

@article{zhang2020physics,
  title={Physics-guided convolutional neural network (PhyCNN) for data-driven seismic response modeling},
  author={Zhang, Ruiyang and Liu, Yang and Sun, Hao},
  journal={Engineering Structures},
  volume={215},
  pages={110704},
  year={2020},
  publisher={Elsevier}
}

\vfill

\end{document}